%% file: egbib.tex
\documentclass[10pt,twocolumn,letterpaper]{article}

\usepackage[pagenumbers]{cvpr} 

\usepackage{graphicx}
\usepackage{amsmath}
\usepackage{amssymb}
\usepackage{booktabs}
\usepackage{times}
\usepackage{epsfig}
\usepackage{multirow}
\usepackage{tabularx}
\usepackage{booktabs}
\usepackage{xcolor}
\usepackage{pifont}
\usepackage{bbm}
\usepackage{algorithm}
\usepackage{algorithmic}
\newcommand{\cmark}{\ding{51}}%
\newcommand{\xmark}{\ding{55}}%

\usepackage[pagebackref,breaklinks,colorlinks]{hyperref}

\usepackage[capitalize]{cleveref}
\crefname{section}{Sec.}{Secs.}
\Crefname{section}{Section}{Sections}
\Crefname{table}{Table}{Tables}
\crefname{table}{Tab.}{Tabs.}

\def\cvprPaperID{459} 
\def\confName{CVPR}
\def\confYear{2022}

\begin{document}

\title{SegTAD: Precise Temporal Action Detection via Semantic Segmentation}

\author{Chen Zhao \quad Merey Ramazanova \quad Mengmeng Xu \quad Bernard Ghanem \\ 
King Abdullah University of Science and Technology (KAUST), Thuwal, Saudi Arabia\\
\tt\small{\{chen.zhao, merey.ramazanova, Mengmeng.Xu, bernard.ghanem\}@kaust.edu.sa}
}

\maketitle

\input{Sections/0_Abstract}
\input{Sections/1_Introduction}

\input{Sections/2_Related_work}
\input{Sections/3_Method}

\input{Sections/4_Experiments}

\input{Sections/5_Conclusion}

{\small
\bibliographystyle{ieee_fullname}
\bibliography{egbib}
}

\end{document}

%% file: Sections/0_Abstract.tex
\begin{abstract}

\begin{figure*} 
\begin{center}
\includegraphics[width=0.97\textwidth]{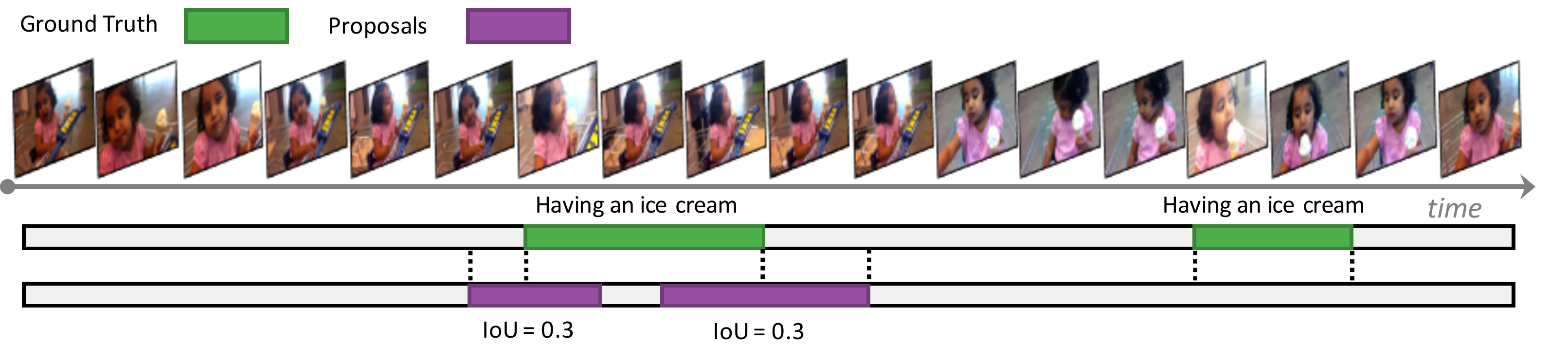}
\end{center}
\vspace{-16pt}
\caption{\textbf{Proposal Annotations.} Proposals different in locations, lengths, and content are assigned the same label if they have the same IoU. These assignments might introduce noise. }
\label{fig:teaser}
\vspace{-12pt}
\end{figure*}
Temporal action detection (TAD) is an important yet challenging task in video analysis. Most existing works draw inspiration from image object detection and tend to reformulate it as a proposal generation - classification problem. However, there are two caveats with this paradigm. \emph{First}, proposals are not equipped  with annotated labels, which have to be empirically compiled, thus  the information in the annotations is not necessarily precisely employed in the model training process. \emph{Second}, there are large variations in the temporal scale of actions, and neglecting this fact may lead to deficient representation in the video features. To address these issues and \emph{precisely} model temporal action detection, we formulate the task of temporal action detection in a novel perspective of semantic segmentation. Owing to the 1-dimensional property of TAD, we are able to convert the coarse-grained detection annotations to fine-grained semantic segmentation annotations for free. We take advantage of them to provide precise supervision so as to mitigate the impact induced by the imprecise proposal labels. We propose an end-to-end framework SegTAD composed of a 1D semantic segmentation network (1D-SSN) and a proposal detection network (PDN). 
We evaluate SegTAD on two important large-scale datasets for action detection and it shows competitive performance on both datasets. 
\vspace{-18pt}

\end{abstract}

%% file: Sections/1_Introduction.tex
\section{Introduction}
Nowadays, millions of videos are produced every day, and high demand arises for automatic video processing and analysis. To this end, various tasks have emerged, for example, action recognition~\cite{8955791}, spatial-temporal action detection~\cite{7044590}, temporal action localization~\cite{8767947, DBLP:journals/tmm/SuZLLH21}. Among those tasks, temporal action detection in untrimmed videos, in particular, is one of the fundamental yet challenging tasks. It requires not only to recognize what actions take place in a video but also to localize when they start and end. 

Most recent works in the literature regard this task as a temporal version of object detection and tackle it by adapting the 2-dimensional solutions on images (e.g., Faster R-CNN \cite{Ren2015FasterRT}) to the 1-dimensional temporal domain for videos \cite{Lin2017SingleST, Chao2018RethinkingTF, Xu2017RC3DRC, Heilbron2016FastTA}. A conventional pipeline is to first identify candidate action segments (i.e., proposals) by analyzing the entire video sequence and then learn to score each segment with an empirically compiled label for each proposal.  This object-detection inspired framework has brought significant improvement on the action detection performance \cite{Heilbron2016FastTA}, especially {with the aid} of deep neural network in recent years \cite{Xu2017RC3DRC, Chao2018RethinkingTF, xu2019g, liu2020multi}. However, it lays two caveats that might lead to imprecise action detection modeling.



\textit{First}, proposals are not accompanied by any annotated labels from the dataset since they are generated on the fly. Their training labels have to be manually compiled based on the ground-truth \emph{action} annotations, i.e.,  the start/end timestamps of \emph{actions} in each video and their corresponding categories. A common practice is to compare each proposal to each ground-truth action in the video in terms of some metric (e.g., temporal Intersection over Union) and use a preset threshold to determine whether a proposal is positive or negative with respect to each category. However, this is obviously not an optimal approach considering that the mapping between action annotations and proposal annotations are not bijective {(as shown in Fig. \ref{fig:teaser})}. Noise is inevitably introduced to the compiled proposal labels regardless of what metric or threshold is adopted, resulting in imprecise modeling.
Note that even in object detection, it is still an open question on how to identify positive and negative proposals, which is crucial to detection performance \cite{Zhang2019BridgingTG}. 

\textit{Second}, object detection is a relatively coarse-grained problem that does not identify every single pixel but predicts a rectangular box surrounding an entire object. However, videos especially in large-scale datasets, e.g., ActivityNet \cite{caba2015activitynet} and HACS \cite{zhao2019hacs}, contain  actions of dramatically varied temporal duration - from less than a second to minutes. Therefore, shifting from the image domain to the video domain without adapting to the video diversity could lead to deficient feature representation {(e.g., burying short actions and under-representing long actions by imprecisely modeling temporal correlations)}, as well as misalignment between proposals and their receptive fields \cite{Chao2018RethinkingTF}. 

To address these issues, in this paper, we propose to formulate the task of temporal action detection (TAD) in a novel perspective with semantic segmentation.  In the image 2-dimensional (2D) domain,  much more effort is demanded to obtain finer-grained annotations for the tasks such as semantic segmentation, considering that not all pixels in a detection bounding box are contained in the object. In contrast, the task of video TAD requires only 1-dimensional (1D) localization of actions --- along the temporal domain. Therefore, all frames within the action boundaries naturally belong to the action category. The detection annotations can be bijectively transformed to segmentation labels without extra effort. We propose an end-to-end temporal action detection framework  to take advantage of the fine-grained prediction of semantic segmentation for more precise detection, dubbed as SegTAD.  SegTAD contains a 1D semantic segmentation network to learn the category of each single frame using the segmentation labels, which are directly transformed from the detection annotations without introducing any label compilation noise.  Regarding the second issue,  we design SegTAD modules based on atrous and graph convolutions to precisely represent actions of various temporal duration.  
{\textbf{The main contributions are:}}
\noindent \textbf{1)} We formulate the task of temporal action detection (TAD) in a novel perspective of semantic segmentation and propose an end-to-end TAD framework SegTAD, which is composed of a 1D semantic segmentation network (1D-SSN) and a proposal detection network (PDN).


\noindent \textbf{2)} In 1D-SSN, we design an hourglass architecture with a module of parallel astrous and graph convolutions to effectively aggregate global features and  multi-scale local features. In PDN, we incorporate a proposal graph  to exploit cross-proposal correlations in our end-to-end framework.


\noindent \textbf{3)} The proposed SegTAD achieves competitive performance on two representative large-scale datasets ActivityNet-v1.3~\cite{caba2015activitynet}  and HACS-v1.1~\cite{zhao2019hacs}.








%% file: Sections/2_Related_work.tex
\section{Related Work}
\subsection{Temporal Action Detection}
Concurrent temporal action detection methods tend to adopt the two-stage framework: 1) generating candidate action segments (i.e., proposals) from the video sequence; 2) cropping each proposal out of the sequence, and classifying each individual proposal to obtain its confidence score. A large number of these methods focus on improving the first stage to generate proposals with high recall, applying an off-the-shelf classifier (e.g., SVM) to get the detection results \cite{Buch2017SSTST, Heilbron2016FastTA, Escorcia2016DAPsDA, Liu2019MultiGranularityGF, Gao2018CTAPCT, zhao2021video}.
Some other methods focus on the second stage, seeking to build more accurate classifiers on proposals produced by other proposal methods (e.g., sliding windows, the above-mentioned first-stage methods) \cite{Shou2016TemporalAL, Shou2017CDCCN, Zeng2019GraphCN, Zhao2017TemporalAD, Ramazanova2022OWLW}. The third category of methods propose end-to-end approaches, where the features of different frames are aggregated and the actions are predicted from the same network~\cite{Heilbron2017SCCSC, Chao2018RethinkingTF, Xu2017RC3DRC, Yuan2017TemporalAL, xu2019g, Long2019GaussianTA, Lin2018BSNBS, Lin2019BMNBN}.
Our paper belongs to the third category. Among these methods, our SegTAD is related to but essentially different from  them  in the following aspects. 

\textbf{Snippet-level classification.} Multiple methods have identified the coarse granularity and regular distribution issues of anchor-based proposals, such as  BSN~\cite{Lin2018BSNBS}, TAG~\cite{Zhao2017TemporalAD},  MGG~\cite{Liu2019MultiGranularityGF}, and CTAP~\cite{Gao2018CTAPCT}. They  have  proposed to incorporate snippet-level proposals as a supplement or replacement to the anchor-based ones. They learn a binary classifier for each snippet, either by a 2D convolutional neural network (CNN) on each snippet~\cite{Gao2018CTAPCT, Zhao2017TemporalAD} or applying a temporal CNN on the entire sequence~\cite{Liu2019MultiGranularityGF, Lin2018BSNBS}. By this means, they obtain the probability of being an action/start/end for each snippet, based on which to generate proposals with flexible duration. In addition, the second-stage method CDC~\cite{Shou2017CDCCN} also classifies each snippet, but the purpose is to refine an existing proposal instead. In this paper, the proposed SegTAD directly formulates  a 1D semantic segmentation problem to classify every single frame into different action categories. It enables the use of large temporal resolution and supports multi-scale feature aggregation with the proposed PAG module. Moreover, it doesn't rely on the actionness/startness/endness scores to generate proposals as in TAG or BSN.


\textbf{Snippet-and-snippet correlations.} The method G-TAD~\cite{xu2019g} exploits temporal correlations between snippets by adopting a graph convolutional network (GCN). It supports limited temporal resolution due to its lack of multi-scale design and the complexity constraint of GCN when more frames are utilized,  consequently sacrificing actions of short duration. Comparatively, our SegTAD adopts an hourglass architecture with an encoder and decoder, and only apply graph convolutions in the intermediate layer with the smallest resolution. In this way, it aggregates global information while preserving the temporal resolution.

\textbf{Proposal-and-proposal correlations.}  BMN~\cite{Lin2019BMNBN} constructs a boundary map with densely-distributed proposals and apply convolutions on the map to utilize the correlations between proposals, whereas 2D-TAN~\cite{zhang2019learning} presents a sparse 2D temporal feature map to represent and correlate proposals. The second-stage method P-GCN~\cite{Zeng2019GraphCN} uses GCNs~\cite{9000721} on proposals obtained by other methods to improve the proposal scores and boundaries. Our SegTAD incorporates graph and edge convolutions to our \emph{end-to-end} detection framework to exploit cross-proposal correlations. Compared to the standalone P-GCN~\cite{Zeng2019GraphCN}, which is essentially a proposal post-processing method and does not consider correlations between frames, SegTAD jointly learns the graph network with the 1D semantic segmentation network and enhances feature representations via cross-frame and cross-proposal aggregation.












    





\begin{figure*} 
\begin{center}
\includegraphics[trim={1.5cm 3.3cm 1.5cm 3cm},width=\textwidth,clip]{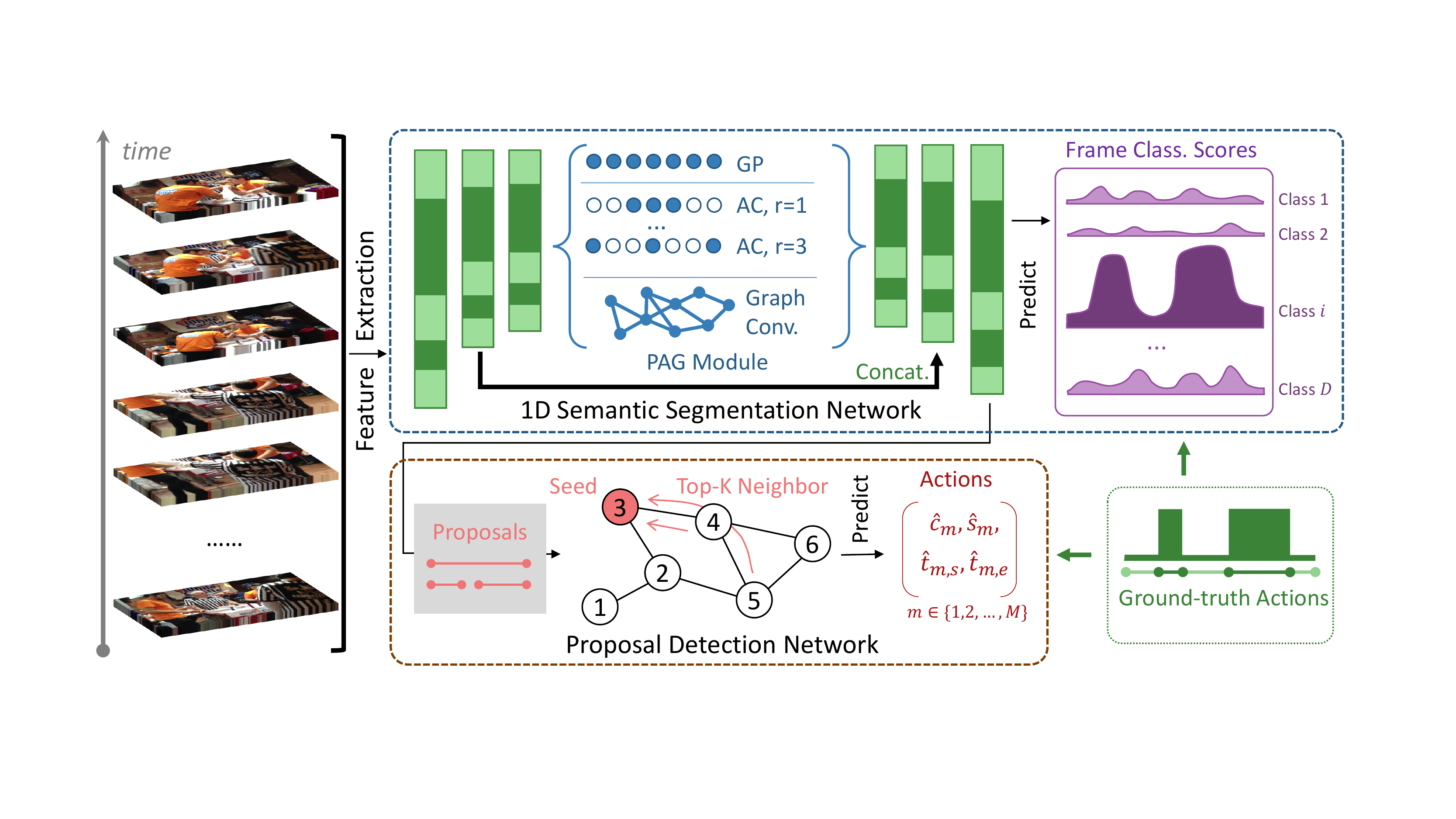}
\end{center}
\vspace{-14pt}
\caption{\textbf{Illustration of our proposed SegTAD architecture.} 
\textbf{Input:} a sequence of video frames; \textbf{Output:} scored candidate actions. \textbf{Top:} 1D Semantic segmentation network (1D-SSN) that learns to classify each frame in the sequence. We design a module of parallel atrous and graph convolutions (PAG) to  effectively aggregate global features and multi-scale local features. \textbf{Bottom:} Proposal detection network (PDN) that scores each candidate action. Graph convolutions are utilized to exploit correlations between proposals.}
\label{fig:arch}
\vspace{-12pt}
\end{figure*}

\subsection{Object Detection and Semantic Segmentation}
In the image domain, object detection~\cite{Girshick2014RichFH, Girshick2015FastR, Ren2015FasterRT, 8979187, 8712433} is a coarse-grained prediction problem, whose output is a rectangular bounding box that surrounds an object in the image. A widely adopted framework for tackling this task is the two-stage method (e.g., R-CNN \cite{Girshick2014RichFH}, Fast R-CNN \cite{Girshick2015FastR}, and Faster R-CNN \cite{Ren2015FasterRT}), which first generates candidate proposals from the original image, and then runs a classifier for each proposal. Recent temporal action detection methods tend to draw inspiration from these object detection methods. But by noticing the 1-dimensional property of the temporal action detection problem, we see that besides object detection, temporal action detection has another analogy in the image domain, which is semantic segmentation.  

Semantic segmentation~\cite{Ronneberger2015UNetCN, Long2015FullyCN, Chen2017RethinkingAC, 8449973, 8269325} is a fine-grained prediction task that predicts the class label of every pixel in an image. Thus, annotating for segmentation usually requires extraordinarily more efforts than for object detection. Compared to object detection which resorts to proposals, semantic segmentation usually adopts a different framework, which seeks to preserve the dense grid of the input image while learning its high-level semantic features with a convolutional network. 
Representative works are U-Net \cite{Ronneberger2015UNetCN} and FCN \cite{Long2015FullyCN}, etc. In videos, temporal semantic annotations and temporal action detection annotations  are bijectively transferable, so no extra efforts are required to annotate segmentation. In this work, taking advantage of this discovery, we utilize the semantic segmentation methodology to formulate the task of temporal action detection. 


%% file: Sections/3_Method.tex
\section{Proposed SegTAD}
\subsection{Problem Formulation and SegTAD Framework}
The task of temporal action detection (TAD) is to predict a set of actions $\Phi=  \left \{ \phi _m=\left ({t}_{m, s},{t}_{m, e}, {c}_m, {s}_m \right ) \right \}_{m=1}^{M}$ given a sequence of $T$ video frames $\{I_t \in \mathbb{R}^{3\times H \times W}\}_{t=1}^T$, where ${t}_{m, s}$ and ${t}_{m, e}$ are  action start and end time respectively, ${c}_m$ is action label, and ${s}_m$ is prediction confidence. To achieve this, we first transform each frame $I_t$ to a 1-dimensional feature vector $\mathbf{x}_t \in \mathbb{R}^{C}$ via a feature extraction network (see Sec.~\ref{sec:exp_setup}). Using the 1D features $\mathbf{x}_t$ as input, we apply our proposed 1D semantic segmentation network (1D-SSN), followed by a proposal detection network (PDN).  The 1D-SSN temporally aggregates features to learn to segment the video sequence in the frame level according to the action annotations, and generates semantic features $\mathbf{y}_t \in \mathbb{R}^{C'}$ for each frame. The PDN learns to score each candidate action and further refines its boundaries. The two components 1D-SSN and PDN are trained end-to-end in a unified architecture end to end. 

We illustrate the entire architecture of SegTAD in Fig.~\ref{fig:arch}. It shows two main components: 1D semantic segmentation network and proposal detection network, which will be described in the following subsections, respectively.




\subsection{1D Semantic Segmentation Network}
Different from conventional TAD works, which perform prediction in the coarse segment level and compile segment labels from  ground-truth action annotations, we  use 1D semantic segmentation (1D-SSN) to learn to predict for each single frame. Based on our 1D-SSN, we are able to take advantage of the original true action annotations without introducing any label noise. In the following,  we first describe the 1D-SSN architecture that aggregates features from a global temporal range as well as multi-scale local range. Then, we present our segmentation loss that uses the original action annotations  to train the segmentation network.
\vspace{-14pt}

\subsubsection{1D-SSN Architecture}
\vspace{-2pt}
Sufficient semantic information from a long temporal range is essential for the task of temporal action detection. This is usually achieved by enlarging the receptive field  via strided 1D convolution or pooling. However, aggressively using these operations will dramatically reduce the temporal resolution and severely impair the feature representation of short actions as a consequence. 

To achieve a large receptive field without severely sacrificing temporal resolution, we consider feature aggregation in two aspects: local feature aggregation and global feature aggregation.  The former aggregates features in a surrounding temporal window to learn local patterns. We need to make it scale-invariant to represent actions of different duration. The latter associates features in a global range, not constrained in the neighborhood of each frame. This breaks the constraints of the temporal locations of each frame and makes use the correlations between frames in the global context~\cite{xu2019g}. 

For the two aspects, we design an hourglass architecture with atrous and graph convolutions in our 1D-SSN. It has the shape of an hourglass, containing an encoder, a parallel module of atrous and graph convolutions (PAG), and a decoder.  The encoder temporally downscales the input features by a small ratio, and the decoder is to restore the temporal resolution.  The PAG module enables local feature aggregation in multiple scales using atrous convolutions and global feature aggregation via graph convolutions. 
\vspace{-10pt}

\subsubsection{1D-SSN Details}
\vspace{-2pt}
\textbf{The encoder}  is comprised of a stack of $L$  strided 1D convolution layers $\mathrm{Conv1D}$$($$k$$=$$3$$,$$s$$=$$2)$, where $k$ is the kernel size, $s$ is the stride, followed by the non-linear activation function  $\mathrm{ReLU}$. We only have $L$$=$$3$ such layers in order not to overly downscale the features.  It applies along the temporal dimension on the input video feature sequence  $\mathbf{X}$$=$$[\mathbf{x}_1, \mathbf{x}_2, \ldots, \mathbf{x}_T]$$\in$$\mathbb{R}^{C\times T}$, where $C$ is the input feature dimension, and transforms it into a representation with lower temporal resolution  $\mathbf{X'}$$=$$[\mathbf{x'}_1, \mathbf{x'}_2, \ldots, \mathbf{x'}_{T/{2^L}}]$$\in$$\mathbb{R}^{C'\times T/{2^L}}$, where $C'$ is the new feature dimension. The encoder reduces the sequence temporal resolution by a factor of ${2^L}$ so as to reduce computation for the subsequent layers, as well as to progressively increase the size of temporal receptive field.

\textbf{The decoder} upscales the temporal resolution of the features $\mathbf{X}'' = \textrm{PAG} (\mathbf{X}')$, where $\textrm{PAG}$ stands for operations in the module of parallel atrous and graph convolutions (PAG) (detailed in following paragraphs). It contains a layer of  linear interpolation to rescale the features along the temporal dimension to the orignal resolution.  In order to complement the details information lost from the encoder, we add a highway connection from the low-level features at the second $\mathrm{Conv1D}$ layer in the encoder, which consists of  $\mathrm{Conv1D} (k=1,  s=1)$, batch normalization and ReLU. Then we concatenate the output of this connection with the interpolated features, and apply a  $\mathrm{Conv1D} (k=3,  s=1)$ layer to adaptively fuse them. With this hourglass (encoder-decoder) architecture, we gradually aggregate features from frames further apart while preserving the temporal resolution of the sequences.

\textbf{The module of parallel atrous and graph convolutions (PAG)} (Fig. \ref{fig:arch}) takes the encoded features $\mathbf{X}'$ as input, and further enlarges the receptive field and  empower the features with scale-invariant capability. Considering that a video sequence usually contains actions of various temporal duration, ranging from a couple of seconds to minutes. Excessive pooling or using strided convolutions could impair short actions as mentioned above, whereas long actions require large receptive field to be semantically represented. To adapt to actions of variant temporal scales, we propose this PAG module, which contains atrous convolutions to aggregate features from multi-scale local neighborhood, and graph convolutions to aggregate features from global context. Note that unlike G-TAD~\cite{xu2019g}, which applies graph convolutions in every layer and consequently incurs huge computation cost, we only have them in this intermediate module after the resolution is reduced by the encoder. In this way, it aggregates global information while preserving the temporal resolution.

\textbf{Atrous convolutions} systematically aggregate multi-scale contextual information without losing resolution. They are able to support expansion of the receptive field~\cite{yu2015multi} by filling in empty elements in the convolutional filter. Compared to normal convolutions, they are equipped with a dilation ratio $d$ to specify the number of empty elements in the filter, reflecting the expansion ratio of the receptive field. We use 4 parallel branches of 1D atrous convolutions with different dilation ratios. {The choice of dilation ratios will be discussed in Sec.~\ref{ablation}}. 




\textbf{Graph convolutions} model the correlations among snippets in a non-local context. We design a graph convolutional network in parallel with the multiple branches of atrous convolutions. Specifically, based on the output features from the encoder $\{\mathbf{x}'_t\}_{t=1}^{T/{2^L}}$ (we call $\mathbf{x}'_t$ features of a snippet in the following), we build a graph denoted as $\mathcal{G}_s=\{\mathcal{V}_s,\mathcal{E}_s\}$.  $\mathcal{V}_s=\{v_t\}_{t=1}^{T/{2^L}}$ refers to the graph nodes, each corresponding to a snippet, and $\mathcal{E}$ denotes graph edges, which represent the correlations between snippets.
To model the correlations of snippets in a global context, we construct the edges  dynamically \cite{Wang2019DynamicGC} according to the semantic similarity between encoded snippet features rather than their temporal locations, which are computed as minus mean square error (MSE) between two feature vectors. If a snippet is among the top $K$ nearest neighbors of another snippet in terms of their semantic similarity, there is an edge connecting them.


With this graph, we apply one layer of edge convolutions to aggregate features of connected nodes \cite{Wang2019DynamicGC}, formulated as  
\begin{equation}
\label{eq:graph_conv}
     \mathbf{X_{GC}}=([\mathbf{X'}^T, \; \mathbf{A}\mathbf{X'}^T-\mathbf{X'}^T] \mathbf{W})^T,
\end{equation}
where $\mathbf{A} \in \mathbb{R}^{T/2^L \times T/2^L}$ is the adjacency matrix defined by edge connections between snippets,  its $(i,j)^\textrm{th}$ element $\mathbf{a}_{i,j} = 1$ if there is an edge between the $i^{\textrm{th}}$ and the $j^\textrm{th}$ snippets, and $\mathbf{a}_{i,j} = 0$, otherwise. For each snippet, $\mathbf{A}\mathbf{X'}^T$ aggregates features from all its connected snippets in the whole sequence. The operation $[\cdot,\cdot]$ concatenates the two feature vectors. $\mathbf{W}$$\in$$\mathbb{R}^{C' \times C'}$ denotes trainable parameters. 
In order to aggregate the global context information along the temporal dimension, we add a global fast path that first does global average pooling and  linearly upsamples back to the original resolution. This mitigates the weight validity issue when large dilation ratios are used \cite{Chen2017RethinkingAC}. 
Then we concatenate the output of all atrous convolutional (AC) branches and the graph convolution (GC) network as well as the global fast path (GP), formulated as 
\begin{equation}
   \textrm{PAG}(\mathbf{X}') = [ \mathbf{X}_{\mathrm{GC}}, \mathbf{X}^1_{\mathrm{AC}}, \ldots, \mathbf{X}^B_{\mathrm{AC}}, \mathbf{X}_{\mathrm{GP}} ].
\end{equation}
Finally, we apply a $\mathrm{Conv1D} \,(k=1, s=1)$ layer followed by ReLU to fuse all branches.
\vspace{-10pt}

\subsubsection{ Segmentation Loss}
\vspace{-2pt}

In 1D-SSN, we formulate the temporal action detection task as a semantic segmentation problem, and predict the category of each single frame to meet their true categories. In the following, we describe how to generate predictions, and  formulate the segmentation loss using  action annotations.



The output from the decoder in 1D-SSN is a sequence of aggregated feature vectors $\mathbf{Y} = [\mathbf{y}_1, \mathbf{y}_2, \ldots, \mathbf{y}_{T}] \in \mathbb{R}^{C'\times T}$. We use this to predict per-frame classification labels. Suppose we have $D$ action categories, applying one layer of linear transformation and Softmax operation yields
\begin{equation}
    \mathbf{p}_t = \textrm{Softmax}(\mathbf{W}_{seg}^T \, \mathbf{y}_t),
\end{equation}
where $\mathbf{p}_t \in \mathbb{R}^{D}$ refers to the predicted label for $t^{\text{th}}$ frame, $\mathbf{W}_{seg} \in \mathbb{R}^{ C' \times D}$ contains the parameters in the linear layer.

If we know the ground-truth label $b_t \in \{1, 2,3,\ldots, D\}$ of each frame, then we can compute the segmentation loss using cross-entropy formulated as follows
\begin{equation}
    \mathcal{L}_{seg} = - \frac{1}{T}\sum _{1\leq t \leq T} \sum _{1\leq d \leq D} \beta_{t,d} \log {p}_{t,d},
\end{equation}
where $\beta_{t,d} = {\mathbbm{1}}\{b_t = d\}$ is the $d^{th}$ one-hot encoded value of the label for the $t^{\text{th}}$ frame.

Now the question becomes how we obtain the segmentation label $b_t$. Assume that a video sequence is annotated with $N$ actions $\Psi  = \left \{ \psi  _n=\left (t_{n, s},t_{n, e}, c_n \right ) \right \}_{n=1}^{N}$, where $t_{n,s}$ and $t_{n,e}$ denote the start and end time of the $n^{\textrm{th}}$ action instance, respectively, and $c_n$ represents its category.  This is a segment-level annotation, without specifying the exact labels of each frame. However, due to the 1D characteristic of the temporal action detection task, we can easily transform this segment-level annotations to finer-grained frame-level labels. 
It assigns a frame the label of an action if it falls inside the action boundaries, otherwise, the frame is labeled as background.  We see that in the entire process, there is no hyper-parameter and the mapping is bijective, which guarantees precise annotation transformation.

Additionally, considering that action boundaries are important for localize an action, we introduce an auxiliary loss 
\begin{multline}
\label{eq:boundary_loss}
    \mathcal{L}_{aux} = \frac{-\sum _{1\leq t \leq T} \beta^s_{t} \log {p}^s_{t} + (1-\beta^s_{t})\log (1-{p}^s_{t})}{T} \\
    =\frac{-\sum _{1\leq t \leq T} \beta^e_{t} \log {p}^e_{t} + (1-\beta^e_{t})\log (1-{p}^e_{t})}{T},
\end{multline}
\noindent where $\beta^s_{t}, \beta^e_{t} \in \{0,1\}$ are start and end labels that indicate whether a frame is the first or last frame of an action. ${p}^s_{t}$ and ${p}^e_{t}$ are predicted confidence scores of a frame being start and end of action, which are generated by 
\begin{equation}
    {\mathbf{p}}^s = \textrm{Sigmoid}(\mathbf{w}_{s2}^T \textrm{ReLU}(\textrm{Conv1d}_{k,s=3,1} (\mathbf{Y}; \mathbf{W}_{s1})))
\end{equation} 
\begin{equation}
    {\mathbf{p}}^e = \textrm{Sigmoid}(\mathbf{w}_{e2}^T \textrm{ReLU}(\textrm{Conv1d}_{k,s=3,1} (\mathbf{Y}; \mathbf{W}_{e1})))
\end{equation} 
\noindent where $\mathbf{W}_{s1}$ and $\mathbf{W}_{e1}$ represent convolutional kernels, and $\mathbf{w}_{s2}, \mathbf{w}_{e2} \in \mathbb{R}^{C'\times1}$ are parameters of the linear layers.

\subsection{Proposal Detection Network}

Considering that the actions in the desired format of $\Phi=  \left \{ \phi _m=\left ({t}_{m, s},{t}_{m, e}, {c}_m, {s}_m \right ) \right \}_{m=1}^{M}$ are not predicted directly by 1D-SSN, we need an extra detection head, for which we design a proposal detection network (PDN). PDN takes the output features from 1D-SSN along with our designed sparse segment patterns as input, and generate predicted actions. This PDN takes advantage of cross-proposal correlations via a graph network, and further enhances the representation of each frame and each proposal.

\noindent \textbf{PDN Architecture:} As shown in Fig.~\ref{fig:arch}, PDN takes the output features $\mathbf{Y}$ from  1D-SSN as well as a sparse pattern of segments. In our framework, in order to precisely detection short actions, we use a high temporal resolution $L=1000$. Therefore, it is cumbersome to enumerate all possible pairs of frames as proposals as done in \cite{xu2019g} and \cite{Lin2019BMNBN}. Instead, we design a sparse pattern of segments, which covers a large variety of action duration and reduces computation compared to dense segments. Let each element $u_{i,j} \in \{0, 1\}$ of the matrix $\mathbf{U} \in \mathbb{R}^{L\times L}$ denote whether the segment starting from $i^{th}$ frame and composed of $j$ frames is selected as a proposal. Its value is determined by the following equation
\begin{equation}
    u_{i,j}= 
    \begin{cases}
    1, &\text{if  $i \, \% \,\eta=0$   and  $j \, \% \, \eta=0$};  \\
    0, &\text{otherwise}.
    \end{cases}
\end{equation}
where $\eta$$=$$8$ is a step size controlling the sparsity degree. With $\Phi$$=$$\{\phi_m$$=$$(t_{m,s}, t_{m,e})\}_{m=1}^M$ being all $M$ proposals specified by $\mathbf{U}$, their features $\mathbf{D} $$=$$\{\mathbf{d}_m$$\in$$ \mathbb{R}^{C'}\}_{m=1}^M$  are extracted  from  video features $\mathbf{Y}$ based on SGAlign~\cite{xu2019g}.

The proposals in the same video are highly correlated and utilizing this property can enhance proposal representations~\cite{Lin2019BMNBN}.  To model correlations between proposals from any temporal locations, we build a second graph $\mathcal{G}_p=\{\mathcal{V}_p,\mathcal{E}_p\}$ on the proposals and take advantage of  graph convolutions in the detection network. Different from the graph $\mathcal{G}_s$ in 1D-SSN, each node in $\mathcal{G}_p$ refers to one proposal, and the edges represent correlations between proposals. Another difference is that the edges $\mathcal{E}_p$ here are constructed based on the temporal intersection over union (tIoU) between proposals, as opposed to the dynamically determined edges in 1D-SSN. We apply the same edge convolutions as shown in Eq.~(\ref{eq:graph_conv}), but define each element $a_{i,j}$ in the adjacency matrix $\mathbf{A}$ as an attention value computed as $a_{i,j} = {\mathbf{d}_i^T \mathbf{d}_j} / {|\mathbf{d}_i| \cdot |\mathbf{d}_j|}$ if there is an edge. We stack 3 layers of edge convolutions in PDN.

In order to efficiently train the proposal network as well as to balance the positive and negative samples, we need to sample from our $M$ proposals. Randomly sampling does not guarantee that the sampled proposals have consistent edge connections with each other to form a meaningful graph. So a better strategy is to sample neighborhoods of proposals rather than individual proposals. We adopt the following sampling strategy based on the SAGE method~\cite{hamilton2017inductive}. We first sample a small number of $M_0$ seed proposals, including $M_0/2$ positive and $M_0/2$ negative samples. Then for each seed proposal, we find its top $K$ neighbors based on its tIoU with other proposals, and put them into the sampling list. For each of the $K$ neighboring proposals, we find its top $K$ neighbors from the remaining proposals, and add these $K\times K$ proposals into the sampling list as well. Hereby, in the sampling list, we totally have  $M_0(1+K + K\times K)$ proposals, all of which have their top $K$ neighbors in the list. ${M_0=50}$ and $K=4$ by default. In inference, we use all $M$ proposals without sampling.


\noindent \textbf{Detection Loss:} The PDN enhances the feature representation of each proposal by aggregating different proposals, formulated as $\mathbf{D}' = \text{PDN} (\mathbf{D})$. We predict proposals' confidence of being actions using the following operation
\begin{equation}
    \mathbf{S} = \text{Sigmoid}(\mathbf{W}_{det}^T \, \mathbf{D}'),
\end{equation}
where $\mathbf{W}_{det} \in \mathbb{R}^{C' \times 2}$  contains the parameters in the linear layer to predict the confidence scores. Note that $\mathbf{S} = [\mathbf{s}_1; \mathbf{s}_2]  $ contains two different scores for each proposal, each corresponding to one loss function we define in the following
\begin{equation}
    \mathcal{L}_{det} =  \mathcal{L}_{reg}(\mathbf{h}_{reg}, \, \mathbf{s}_{1}) + \mathcal{L}_{cls}(\mathbf{h}_{cls}, \, \mathbf{s}_{2}),
\end{equation}
where $\mathbf{h}_{reg}$ is the tIoU between each proposals  and their closest ground-truth actions, and $\mathcal{L}_{reg}$ is computed using  mean square errors. $\mathbf{h}_{cls} = \mathbbm{1} (\mathbf{h}_{reg} > \tau)$, where $\tau=0.5$ is an tIoU threshold determining whether a proposal is positive or negative, and $\mathcal{L}_{cls}$ is a binary cross-entropy loss similarly computed as either term in Eq.~(\ref{eq:boundary_loss}).


\begin{table}[t]
\centering
\caption{\textbf{Action detection result comparisons on validation set of ActivityNet-v1.3}, measured by mAP ($\%$) at different tIoU thresholds and the average mAP. G-TAD achieves better performance in average mAP than the other methods, even the latest work of BMN and P-GCN shown in the  second-to-last block. ($^*$ Re-implemented with the same features as ours.) 
}
\vspace{-8pt}
\small
\begin{tabular}{l|ccc|c}
\toprule
\textbf{Method}  & 0.5  &  0.75  & 0.95 & \textbf{Average}\\
\midrule
Wang \textit{et al.} \cite{wang2016uts}    & 43.65 & -  & - & -\\
Singh \textit{et al.} \cite{singh2016untrimmed}  & 34.47 & - & - & - \\
SCC \cite{Heilbron2017SCCSC}   & 40.00 & 17.90  & 4.70   & 21.70  \\
Lin \textit{et al.} \cite{lin2017temporal}  & 44.39   & 29.65  & 7.09  & 29.17  \\
CDC \cite{Shou2017CDCCN}  & 45.30 & 26.00 & 0.20 & 23.80 \\
TCN \cite{dai2017temporal}  & 37.49 & 23.47 & 4.47 & 23.58 \\
R-C3D \cite{Xu2017RC3DRC}   & 26.80 & - & - & - \\
SSN \cite{Zhao2017TemporalAD}  & 34.47 & - & - & - \\
BSN \cite{Lin2018BSNBS}   & 46.45  & 29.96 & 8.02  & 30.03  \\
TAL-Net \cite{Chao2018RethinkingTF}  & 38.23 & 18.30 & 1.30 & 20.22 \\ 
P-GCN+BSN \cite{Zeng2019GraphCN}   &48.26 &33.16 &3.27 &31.11  \\
BMN~\cite{Lin2019BMNBN}   & \underline{50.07} & \textbf{34.78} & {8.29} & \underline{33.85}  \\
BMN$^*$~\cite{Lin2019BMNBN}  & 48.56 &33.66 &\underline{9.06}  &33.16 \\
I.C \& I.C~\cite{Zhao2020BottomUpTA}  &43.47&33.91&\textbf{9.21}&30.12  \\
\hline
SegTAD (top-1 cls.) & {49.86} &{34.37} &6.50 &{33.53}  \\
SegTAD (top-2 cls.) & \textbf{50.52} &\underline{34.76} &6.85 &\textbf{33.99}  \\
\bottomrule
\end{tabular}
\label{tab:sota_anet}
\vspace{-8pt}
\end{table}

\subsection{Training and Inference}
\noindent \textbf{Training:} We train the proposed SegTAD end to end using the segmentation loss and the detection loss, as well as  the auxiliary loss as follows
\begin{equation}
    \mathcal{L} = \mathcal{L}_{seg} + \lambda_1 \mathcal{L}_{det} + \lambda_2 \mathcal{L}_{aux} + \lambda_3 \mathcal{L}_r,
\end{equation}
where $\mathcal{L}_r$ is  $\mathcal{L}_2$-norm, $\lambda_1$, $\lambda_2$, $\lambda_3$ denotes the weights for different loss terms, which are all set to 1 by default.

\noindent \textbf{Inference:} At inference time, we compute the score of each candidate action using the two scores predicted from the proposal detection network as $s_m = s_{m,1} \times s_{m,2}$.   Then we run soft non-maximum suppression (NMS) using the scores and keep the top 100 predicted segments as final output.

\begin{table}[t]
\centering
\caption{\textbf{Action detection results on HACS-v1.1}, measured by mAP ($\%$) at different tIoU thresholds and the average mAP. }
\vspace{-10pt}
\small
\begin{tabular}{l|ccc|c|c}
        \toprule
\textbf{Method} & \multicolumn{4}{c|}{Validation}  & Test \\ 
  & 0.5  &  0.75  & 0.95 & Average & Average\\
\midrule
SSN \cite{zhao2019hacs}  & 28.82 & 18.80  & 5.32 & 18.97 & 16.10 \\
BMN~\cite{HACS2019}  & - & - & - & - & 22.10 \\
S-2D-TAN~\cite{zhang2019learning}  & - & - & - & - & 23.49 \\
SegTAD  & \textbf{43.33} &  \textbf{29.65} & \textbf{6.23}  &  \textbf{29.24} & \textbf{28.90} \\
\bottomrule
\end{tabular}
\label{tab:sota_hacs}
\vspace{-10pt}
\end{table}

%% file: Sections/4_Experiments.tex
\section{Experimental Results}
\subsection{Datasets and Implementation Details} \label{sec:exp_setup}
\textbf{Datasets and Evaluation Metric}.
We conduct our experiments on two large-scale action understanding dataset,
{\bf ActivityNet-v1.3} \cite{caba2015activitynet} and 
\textbf{HACS-v1.1}
\cite{zhao2019hacs} for the temporal action localization task. {\bf ActivityNet-v1.3} contains around $20,000$ temporally annotated untrimmed videos with $200$ action categories. Those videos are randomly divided into training, validation and testing sets by the ratio of 2:1:1. 
\textbf{HACS-v1.1} follows the same annotation scheme as ActivityNet-v1.3. It also includes $200$ action categories but collects $50,000$ untrimmed videos for the temporal action localization task. We evaluate SegTAD performance with the average of mean Average Precision (mAP)  over 10 different IoU thresholds $[0.5\text{:} 0.05\text{:}0.95]$ on both datasets. 

\textbf{Implementations.} For ActivityNet-v1.3, we sample each video at 5 frames per second and adopt the  two-stream network by Xiong~et.~al.~\cite{xiong2016cuhk} pre-trained on Kinetics-400 \cite{quo2017i3d} to extract frame-level features, and rescale each sequence into 1000 snippets as SegTAD input.
For HACS, we use {the publicly available features } extracted using an I3D-50 \cite{quo2017i3d} model pre-trained on Kinetics-400 \cite{quo2017i3d} and temporally rescale them into 400 snippets. 
We implement and test our framework using PyTorch 1.1, Python 3.7, and CUDA 10.0.
In training, the learning rates are $1\text{e}{-5}$ on ActivityNet-1.3 and $2\text{e}{-3} $ on HACS-v1.1 for the first $7$ epochs, and are reduced by $10$ for the following $8$ epochs. 
In inference, we leverage the global video context and take the top-1 or top-2 video classification scores from the action recognition models of \cite{wang2017untrimmed} and \cite{lin2019tsm}, respectively for the two datasets, and multiply them by the confidence score $c_j$ for evaluation.


\begin{figure}
\begin{center}
\includegraphics[width=0.34\textwidth]{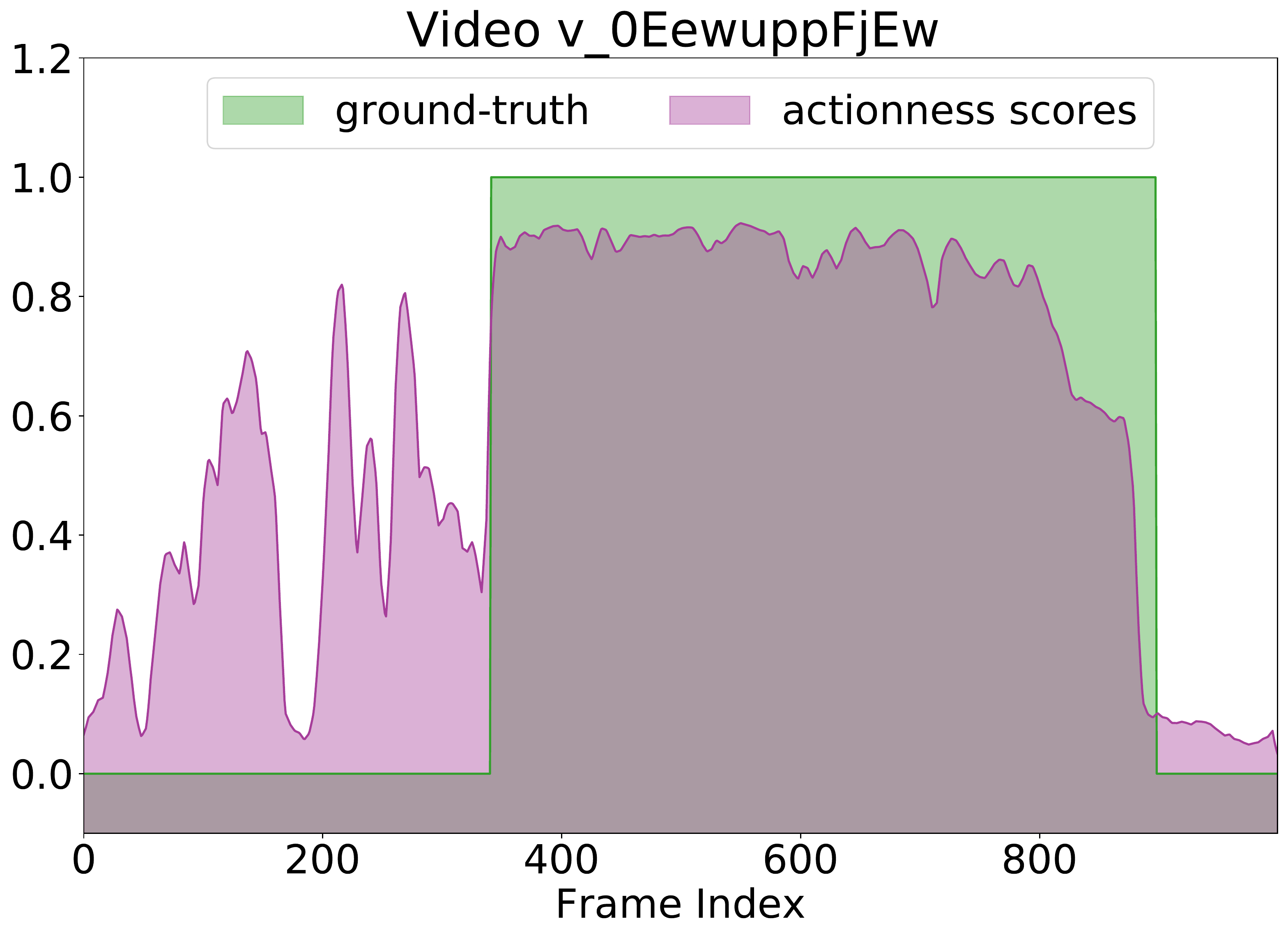} %
\includegraphics[width=0.34\textwidth]{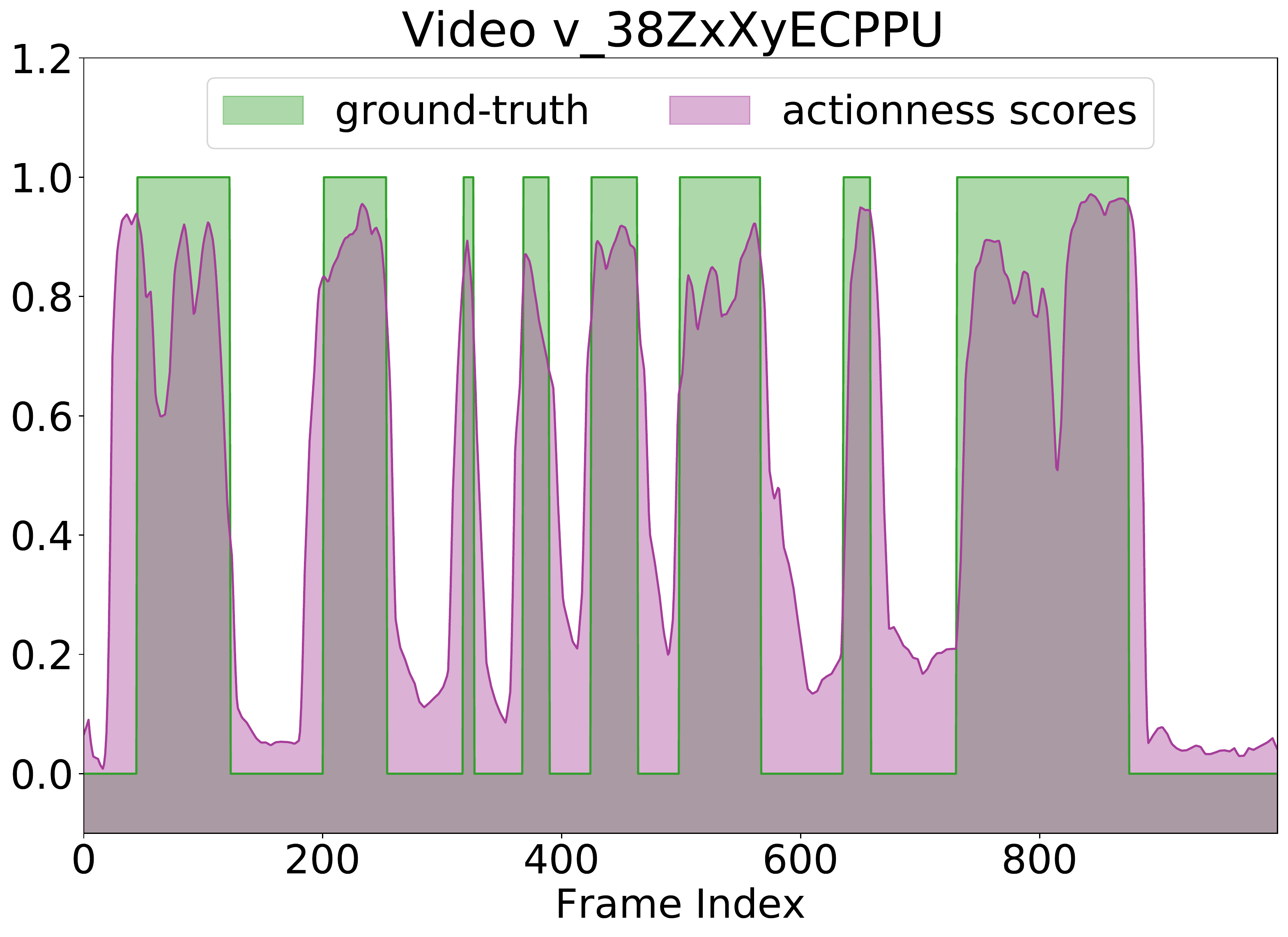}
\end{center}
\vspace{-14pt}
\caption{\textbf{Predicted per-frame classification scores compared to ground-truth labels.} We only plot the scores of the ground-truth category. Green curves represent the ground-truth, and score = 1.0 represents the frames are inside action, and score =0.0 otherwise. Purple curves represent the predicted scores for each frame.}
\label{fig:action}
\vspace{-15pt}
\end{figure}

\subsection{Comparison to State-of-the-Art}
In Table~\ref{tab:sota_anet} and Table~\ref{tab:sota_hacs}, we compare SegTAD with representative temporal action detectors in the literature on the two datasets. We report mAP at different tIoU thresholds, as well as average mAP. 

On ActivityNet-v1.3, SegTAD achieves competitive average mAP of $33.99\%$, significantly outperforming the recent works I.C \& I.C ~\cite{Zhao2020BottomUpTA} and BMN~\cite{Lin2019BMNBN}. Notably, BMN extracts  video features from ActivityNet-finetuned model such that the extracted features are more distinguishable on the target dataset. In contrast, we use more general Kinetics-pretrained features. To achieve fair comparison, we also show the re-produced  BMN experimental results with the same features as ours, and our performance gain  is even more remarkable.
On HACS-v1.1, SegTAD reaches $28.90\%$ average mAP on the test set, surpassing both the challenge winner S-2d-TAN~\cite{zhang2019learning} and BMN~\cite{Lin2019BMNBN} by large margins. Compared with ActivityNet-v1.3, HACS-v1.1 is more challenging because of its substantial data-scale and precise segment annotations.  Therefore, our superior performance on HACS-v1.1 makes SegTAD more remarkable.

\begin{table}[t]
\centering
\small
\caption{\textbf{Effectiveness of our segmentation network.}  
}
\vspace{-8pt}
\small
\begin{tabular}{c|cccc}
\toprule
Segmentation loss   & 0.5  &  0.75  & 0.95  & Avg.   \\
\hline
\xmark & 49.15	&33.45&	3.81 &	32.45\\ 
\cmark& \textbf{49.86} &\textbf{34.37} &\textbf{6.50} &\textbf{33.53} \\
\bottomrule
\end{tabular}
\label{tab:components}
\vspace{-4pt}
\setlength{\belowcaptionskip}{-1cm}
\end{table}

\begin{table}[t]
\small
\centering
\caption{\textbf{Different loss functions for segmentation.} 
}
\vspace{-8pt}
\small
\begin{tabular}{c|cccc}
\toprule

  Segment. loss types& 0.5  &  0.75  & 0.95  & Avg.\\
\hline
{Binary} &49.35 &	33.77	&4.07&	32.79  \\
SegTAD &  \textbf{49.86} &	\textbf{34.37}  &\textbf{6.50}  & \textbf{33.53}\\
\bottomrule
\end{tabular}
\label{tab:seg_loss}
\vspace{-6pt}
\setlength{\belowcaptionskip}{4cm}
\end{table}

\subsection{Ablation Study}
\label{ablation}
In this subsection, we provide ablation study to demonstrate the importance of the proposed 1D semantic segmentation network to the detection performance. Also we verify the effectiveness of our design choice for the 1D semantic segmentation network (1D-SSN) and proposal detection network (PDN).
In Table~\ref{tab:components}, we compare SegTAD to its variants of disabling the segmentation loss in the 1D-SSN component. We can see that using the loss  leads to obvious improvement compared to not using it.
In Table~\ref{tab:seg_loss}, we show the performance of replacing the segmentation loss using a binary classification loss, which learns whether a frame is inside an action or not. Our multi-class segmentation loss in SegTAD is obviously better.
\begin{table}[t]
\centering
\small
\caption{\textbf{Ablating studies in PAG of 1D-SSN.} AC: Atrous convolutions, GC: graph convolutions, GP: global fast path.}
\vspace{-8pt}
\small
\begin{tabular}{ccc|cccc}
\toprule
\multicolumn{3}{c|}{PAG Branches} & \multicolumn{4}{c}{mAP at different tIoUs} \\
\hline
 AC & GC &  GP& 0.5  &  0.75  & 0.95  & Avg.   \\
\hline

\xmark& \cmark &\cmark & 48.55&	33.04	& 5.21	&32.37   \\
 \cmark & \xmark &\cmark& 49.48 &	33.95	&\textbf{7.50}	& 33.30  \\
 \cmark & \cmark &\xmark&49.75&	34.25&	5.97	&33.29 \\ 
 \hline
\cmark &\cmark & \cmark & \textbf{49.86} &	\textbf{34.37}  &{6.50}  & \textbf{33.53} \\
\bottomrule
\end{tabular}
\label{tab:1D_SSN}
\vspace{-4pt}
\end{table}
\begin{table}[t]
\centering
\small
\caption{\textbf{Different sets of dilation ratios of the AC branches.}}
\vspace{-8pt}
\begin{tabular}{c|cccc}
\toprule
Dilation ratios & 0.5  &  0.75  & 0.95  & Avg.  \\
\hline
 1, 2, 4, 6 & 49.58&	34.14&	5.58&	33.25\\ 
1, 6, 12, 18 &49.71&34.01&	5.90	&33.31 \\
1, 10, 20, 30&\textbf{49.86} &	\textbf{34.37}  &\textbf{6.50}  & \textbf{33.53} \\
1, 16, 32, 64 & 50.00&	34.31	&6.35&	33.45 \\
\bottomrule
\end{tabular}
\label{tab:atrous_conv}
\vspace{-14pt}
\end{table}
In 1D-SSN, the module of parallel atrous and graph convolutions (PAG) is important to aggregate features from multiple scales. We ablate different branches in PAG to show the performance change in Table~\ref{tab:1D_SSN}. It shows that the network with all three kinds of branches produce the best performance. 


We also show the results of different sets of dilation ratios for the atrous convolution branches in Table~\ref{tab:atrous_conv} and choose 1, 10, 20, 30 due to its highest mAP. 
In this experiment, we evaluate the effectiveness of our proposal detection network (PDN) by replacing its each layer of edge convolutions  with a layer of  $\text{Conv1D}(k=1, s=1)$  and apply it on each single proposal independently. In this way, this variant cannot make use of the cross-proposal correlations. We can see from Table~\ref{tab:abl_PDN} that using graph convolutions brings significant improvement compared to independently learning for each proposal.
In Table~\ref{tab:node_sim}, we compare different  metrics to determine the similarity between proposals: tIoU and distance between proposal centers. We adopt tIoU in SegTAD due to its better performance.

\begin{table}[t]
\small
\centering
\caption{\textbf{Ablating the proposal detection network.} 
}
\vspace{-8pt}
\small
\begin{tabular}{c|cccc}
\toprule
 PDN layers  & 0.5  &  0.75  & 0.95  & Avg.\\
\hline
 $\text{Conv1D}(k=1, s=1)$  &48.31&	32.79&5.49&	32.12\\ 
Graph convolutions &   \textbf{49.86} &	\textbf{34.37}  &\textbf{6.50}  & \textbf{33.53}\\
\bottomrule
\end{tabular}
\label{tab:abl_PDN}
\vspace{-6pt}
\end{table}

\begin{table}[t]
\centering
\caption{\textbf{Comparing different similarity metrics: temporal intersection over union and center distance between proposals.}  
}
\vspace{-8pt}
\small
\begin{tabular}{c|cccc}
\toprule
Node similarity metric  & 0.5  &  0.75  & 0.95  & Avg.\\
\hline
 Center distance  &  49.34 &	33.57&	3.62	&32.52\\
Temp. intersection over union &  \textbf{49.86} &	\textbf{34.37}  &\textbf{6.50}  & \textbf{33.53}\\
\bottomrule
\end{tabular}
\label{tab:node_sim}
\vspace{-14pt}
\setlength{\belowcaptionskip}{-1cm}
\end{table}


\subsection{Visualization of Segmentation Output}
We visualize the output from the 1D semantic segmentation network and compare to ground-truth labels in Fig.~\ref{fig:action}. Our output tightly matches the ground-truth even for the video that contains many short action instances such as the bottom example. Such accurate segmentation is important for learning distinctive features for each frame, and consequently benefits the final detection.









%% file: Sections/5_Conclusion.tex
\section{Conclusion}
In this paper, we propose a novel perspective to formulate the task of temporal action detection based on 1D semantic segmentation to achieve more accurate label assignment and precise localization. We propose a  temporal action detection framework SegTAD, which is composed of a 1D semantic segmentation network (1D-SSN) and a proposal detection network (PDN). To suit the large variety of action temporal duration, in 1D-SSN, we design a module of parallel atrous and graph  convolutions (PAG) to aggregate multi-scale local features and global features. In PDN, we design a second graph network to model the cross-proposal correlations. SegTAD is an end-to-end framework that is trained jointly using the segmentation and detection losses from both 1D-SSN and PDN, respectively. As a conclusion, we would like to emphasize the need to focus more on the unique characteristics of videos when dealing with detection problems in video.

